\documentclass{article} 
\usepackage[preprint]{colm2026_conference}

\usepackage{microtype}
\usepackage{hyperref}
\usepackage{url}
\usepackage{graphicx}
\usepackage{amsmath}
\usepackage{amssymb}

\usepackage{booktabs}
\usepackage{colortbl}
\usepackage{multirow}
\usepackage{enumitem}
\usepackage{algorithm}
\usepackage{algpseudocode}
\usepackage{tikz}
\usetikzlibrary{arrows.meta,positioning,calc,fit,backgrounds,decorations.pathreplacing,shapes.geometric}

\usepackage{lineno}

\definecolor{darkblue}{rgb}{0, 0, 0.5}
\definecolor{mygray}{gray}{0.92}
\hypersetup{colorlinks=true, citecolor=darkblue, linkcolor=darkblue, urlcolor=darkblue}

\title{Internalizing Outcome Supervision into Process Supervision:\\A New Paradigm for Reasoning Reinforcement Learning}

\author{Fei Ding\thanks{Corresponding author: \texttt{dignfei@gmail.com}} \quad Yongkang Zhang \quad Runhao Liu \\
Alibaba Group \\
\And
Yuhao Liao \quad Zijian Zeng \quad Sibo Wang \quad Huiming Yang \\
Tsinghua University
}

\begin{document}

\ifcolmsubmission
\linenumbers
\fi

\maketitle

\begin{abstract}
A central challenge in reasoning reinforcement learning lies not merely in the sparsity of outcome supervision, but in converting end-of-sequence outcome feedback into fine-grained learning signals that can act on intermediate reasoning steps. Existing approaches either perform sequence-level optimization with outcome rewards, which precludes precise credit assignment, or rely on externally constructed process supervision at substantial cost and limited scalability. We propose a new perspective: reasoning reinforcement learning as the problem of \emph{internalizing outcome supervision into process supervision}. Building on this perspective, we introduce IOP, a framework that enables the model to automatically distill process-level learning signals by identifying, repairing, and reusing failed reasoning trajectories---achieving finer-grained policy optimization under outcome supervision alone. We further formalize this idea as a training paradigm in which the model continuously generates and refines internal process supervision throughout reinforcement learning, offering a new pathway for fine-grained credit assignment that is distinct from exogenous process supervision.
\end{abstract}

\begin{figure}[h]
    \centering
    \resizebox{\columnwidth}{!}{%
    \begin{tikzpicture}[
        >=Stealth,
        tokenstyle/.style={minimum width=0.38cm, minimum height=0.38cm, font=\tiny, inner sep=0.5pt, rounded corners=1pt},
        tok_ok/.style={tokenstyle, fill=green!15, draw=green!40},
        tok_err/.style={tokenstyle, fill=red!20, draw=red!50},
        tok_fix/.style={tokenstyle, fill=orange!25, draw=orange!60},
        gate_on/.style={tokenstyle, fill=yellow!35, draw=yellow!70, font=\tiny\bfseries},
        gate_off/.style={tokenstyle, fill=gray!8, draw=gray!25, text=gray!40, font=\tiny},
        arr/.style={->, semithick, >=Stealth},
        titlefont/.style={font=\scriptsize\bfseries},
        lblfont/.style={font=\tiny},
        ]

        \begin{scope}[on background layer]
            \fill[red!3, draw=red!40, rounded corners=4pt, semithick]
            (-4.85, -1.55) rectangle (-0.35, 1.55);
            \fill[blue!3, draw=blue!40, rounded corners=4pt, semithick]
            (0.35, -1.55) rectangle (6.65, 1.55);
        \end{scope}

        \node[titlefont, text=red!60!black] at (-2.6, 1.25) {(a) Outcome Supervision};

        \node[tok_ok]  (a1) at (-4.2, 0.55) {$z_1$};
        \node[tok_ok,  right=0.5pt of a1] (a2) {$z_2$};
        \node[tok_ok,  right=0.5pt of a2] (a3) {$z_3$};
        \node[tok_err, right=0.5pt of a3] (a4) {$z_4$};
        \node[tok_err, right=0.5pt of a4] (a5) {$z_5$};
        \node[tok_ok,  right=0.5pt of a5] (a6) {$z_6$};
        \node[tok_err, right=0.5pt of a6] (a7) {$z_7$};
        \node[right=2pt of a7, font=\tiny\bfseries, text=red!65!black] {$\boldsymbol{\times}$};

        \foreach \i in {1,...,7} {
            \draw[arr, red!45, thin] ([yshift=-1pt]a\i.south) -- ++(0, -0.35)
            node[below, font=\tiny, text=red!50, inner sep=0pt] {$\nabla^{\!-}$};
        }

        \node[font=\tiny, text=red!55!black] at (-2.6, -1.15)
        {Uniform penalty on \textbf{all} tokens};

        \draw[->, line width=1.5pt, violet!55, shorten >=2pt, shorten <=2pt]
        (-0.2, 0) -- (0.2, 0)
        node[midway, above=2pt, font=\tiny\bfseries, text=violet!65!black] {Internalize};

        \node[titlefont, text=blue!60!black] at (3.5, 1.25) {(b) IOP: Internalized Process Supervision};

        \node[lblfont, text=red!55!black, anchor=east] at (1.85, 0.65) {\textit{Failed}};
        \node[tok_ok]  (b1) at (2.15, 0.65) {$y_1$};
        \node[tok_ok,  right=0.5pt of b1] (b2) {$y_2$};
        \node[tok_ok,  right=0.5pt of b2] (b3) {$y_3$};
        \node[tok_err, right=0.5pt of b3] (b4) {$y_4$};
        \node[tok_err, right=0.5pt of b4] (b5) {$y_5$};
        \node[tok_ok,  right=0.5pt of b5] (b6) {$y_6$};
        \node[tok_err, right=0.5pt of b6] (b7) {$y_7$};
        \node[right=2pt of b7, font=\tiny\bfseries, text=red!65!black] {$\boldsymbol{\times}$};

        \node[lblfont, text=green!45!black, anchor=east] at (1.85, 0.05) {\textit{Repaired}};
        \node[tok_ok]  (c1) at (2.15, 0.05) {${y}_1$};
        \node[tok_ok,  right=0.5pt of c1] (c2) {${y}_2$};
        \node[tok_ok,  right=0.5pt of c2] (c3) {${y}_3$};
        \node[tok_fix, right=0.5pt of c3] (c4) {$\tilde{y}_4$};
        \node[tok_fix, right=0.5pt of c4] (c5) {$\tilde{y}_5$};
        \node[tok_ok,  right=0.5pt of c5] (c6) {${y}_6$};
        \node[tok_fix, right=0.5pt of c6] (c7) {$\tilde{y}_7$};
        \node[right=2pt of c7, font=\tiny\bfseries, text=green!50!black] {$\boldsymbol{\checkmark}$};

        \foreach \i in {4,5,7} {
            \draw[arr, orange!70, thin] (b\i.south) -- (c\i.north);
        }

        \node[lblfont, text=black!55, anchor=east] at (1.85, -0.6) {\textit{Gradient}};
        \node[gate_off] (g1) at (2.15, -0.6)  {\text{--}};
        \node[gate_off, right=0.5pt of g1] (g2) {\text{--}};
        \node[gate_off, right=0.5pt of g2] (g3) {\text{--}};
        \node[gate_on,  right=0.5pt of g3] (g4) {$\nabla$};
        \node[gate_on,  right=0.5pt of g4] (g5) {$\nabla$};
        \node[gate_off, right=0.5pt of g5] (g6) {\text{--}};
        \node[gate_on,  right=0.5pt of g6] (g7) {$\nabla$};

        \draw[arr, blue!50, thin] (c4.south) -- (g4.north);
        \draw[arr, blue!50, thin] (c5.south) -- (g5.north);
        \draw[arr, blue!50, thin] (c7.south) -- (g7.north);

        \node[font=\tiny, text=blue!55!black] at (3.5, -1.15)
        {Targeted gradient \textbf{only} at error tokens};

    \end{tikzpicture}
    }%
    \vspace{-2pt}
    {\scriptsize\textbf{+6.9\%} avg accuracy $\cdot$ \textbf{2.3$\times$} sample efficiency $\cdot$ No external annotations\par}
    \vspace{2pt}
    \caption{IOP turns outcome feedback into process-level supervision by fixing failed trajectories and penalizing only the erroneous tokens.}
    \label{fig:teaser}
\end{figure}
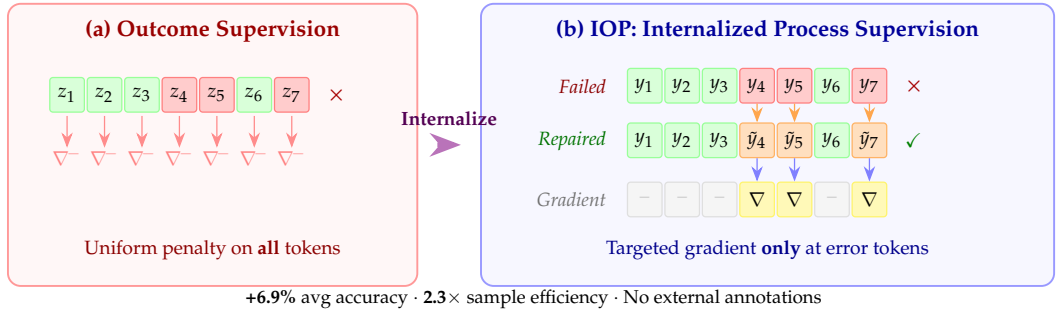

\section{Introduction}
\label{sec:intro}

Recent advances in large language model (LLM) reasoning---spanning mathematics, code generation, and complex logical tasks---have been substantially driven by reinforcement learning (RL). Outcome-based RL has proven effective for improving performance on challenging reasoning benchmarks \citep{havrilla2024teachinglargelanguagemodels,shao2024deepseekmathpushinglimitsmathematical,Guo_2025,openai2024o1}, with methods such as RLVR and GRPO directly optimizing final-answer correctness through verifiable outcome rewards \citep{shao2024deepseekmathpushinglimitsmathematical,wen2026reinforcement}. However, as reasoning chains grow longer, the limitations of outcome-level supervision become increasingly apparent: a failed trajectory is rarely wrong in its entirety---it typically contains many correct intermediate steps alongside a few critical errors. Uniformly rewarding or penalizing the entire trajectory based solely on the final outcome conflates correct local reasoning with the actual sources of failure. The fundamental difficulty in reasoning RL, therefore, is not merely that outcome supervision is sparse, but that it lacks the granularity required for effective credit assignment to intermediate reasoning steps.

To address this, researchers have introduced process supervision that explicitly models intermediate reasoning steps. Seminal work demonstrated that step-by-step verification substantially improves complex mathematical reasoning compared to outcome supervision alone \citep{lightman2024lets}. Subsequent efforts explored automated construction of process supervision---including automatic intermediate-step sampling, training of process reward models (PRMs), and search- or verifier-based identification of critical error steps \citep{luo2024improvemathematicalreasoninglanguage,zhang-etal-2025-lessons,yang-etal-2025-beyond-first}. More recent work further integrates process signals directly into RL through error-region penalties, prefix-level supervision, or localized process optimization \citep{liu2026savegoodprefixprecise,liang2026learningirrecoverableerrorlocalizedpolicy,nie2026attnpoattentionguidedprocesssupervision}. Collectively, these studies establish that process supervision is essential for effective credit assignment in long-chain reasoning.

Yet existing process supervision methods predominantly rely on externally provided signals---whether human step annotations, separately trained PRMs, explicitly constructed verifiers, or search-derived pseudo-labels---all of which constitute \emph{exogenous} process supervision. While effective, such methods incur substantial construction and maintenance costs and are difficult to scale as the policy model improves. In other words, prior work has established \emph{that} process supervision matters, but has not adequately addressed a more fundamental question: \emph{can the model generate its own process supervision during reinforcement learning, given only outcome-level feedback?}

This paper departs from precisely this question and proposes a new perspective: reasoning RL as the problem of \emph{internalizing outcome supervision into process supervision}. The key observation is that failed trajectories are not uniformly negative samples but contain structured information that can be corrected and reused. If the model can generate a repaired version of a failed trajectory and distill critical changes from the difference between the two, outcome supervision can be converted into process supervision that acts on intermediate steps.
This perspective resonates with recent work on self-correction and reflective reasoning, which shows that models can perform self-verification and correction during inference, and that this capability can be strengthened via online RL \citep{kumar2025training,ma-etal-2025-s2r,xiong2025selfrewardingcorrectionmathematicalreasoning,lee2025revise}. We extend this self-correction capability from inference time to training time, thereby proposing a new training paradigm: under outcome supervision alone, the model continuously generates and refines internal process supervision to achieve finer-grained policy optimization.

\paragraph{Contributions.}
\begin{itemize}
    \item We reframe reasoning RL as the problem of internalizing outcome supervision into process supervision, and establish a corresponding training paradigm in which the model automatically distills process-level signals through failure repair during RL.

    \item We propose the IOP framework and its GSPO instantiation IOP-GSPO, which converts sequence-level rewards into token-level gating signals via audit gating, minimum-edit repair, and verification-based adaptive truncation.

    \item Experiments demonstrate a synergistic reinforcement loop between policy and repair capabilities; IOP-GSPO consistently outperforms GSPO (+4.9--6.9\%) and exogenous process supervision methods across three reasoning benchmarks, with approximately $2.3\times$ sample efficiency.
\end{itemize}

\section{Related Work}
\label{sec:related_work}

\paragraph{Outcome-supervised reasoning RL.}
Mainstream open-source reasoning RL methods rely on outcome supervision for sequence-level optimization, including GRPO \citep{shao2024deepseekmathpushinglimitsmathematical}, GSPO \citep{zheng2025groupsequencepolicyoptimization}, DPO \citep{NEURIPS2023_a85b405e}, and the broader RLVR line \citep{havrilla2024teachinglargelanguagemodels,wen2026reinforcement,Guo_2025,openai2024o1}. These methods achieve strong results on math, code, and complex reasoning tasks, yet supervision operates on entire trajectories, leaving local errors in long-chain reasoning unattributable to specific intermediate steps.

\paragraph{Explicit process supervision and process reward models.}
A complementary line of work provides explicit process supervision through intermediate-step verification, scoring, or training of process reward models (PRMs). \citet{lightman2024lets} demonstrated the value of step-level supervision for complex mathematical reasoning; \citet{wang-etal-2024-math,luo2024improvemathematicalreasoninglanguage} further explored automated process label construction, PRM training, and scaling of process data via automated supervision. Recent studies note that reflection and self-correction complicate first-error localization \citep{zhang-etal-2025-lessons,yang-etal-2025-beyond-first}, and generative verifiers with data-efficient process supervision modeling have gained attention \citep{khalifa2026process}. This line establishes the importance of process supervision for fine-grained attribution, but the signal sources remain limited to human annotations, external PRMs, or static labels.

\paragraph{Process-supervised RL and fine-grained attribution.}
More recent work integrates process supervision directly into RL rather than using it solely as an independent verifier---for example, through prefix separation, error-region penalties, or attention-guided updates to improve policy update precision \citep{liu2026savegoodprefixprecise,yao2026prlprocessrewardlearning,pronesti2026outcomeverificationverifiableprocess,liang2026learningirrecoverableerrorlocalizedpolicy,nie2026attnpoattentionguidedprocesssupervision}. These methods demonstrate that process-level signals can directly improve local update quality, yet they still rely on explicit error boundaries, step labels, or independent PRMs---the signal source remains fundamentally exogenous.

\paragraph{Self-correction, self-verification, and repair-based reasoning.}
Another body of work investigates whether models can improve reasoning through correction and verification, including Self-Refine \citep{NEURIPS2023_91edff07}, Reflexion \citep{NEURIPS2023_1b44b878}, SCoRe \citep{kumar2025training}, S$^2$R \citep{ma-etal-2025-s2r}, and self-rewarding correction \citep{xiong2025selfrewardingcorrectionmathematicalreasoning}. These methods show that models can generate valuable internal feedback to improve output quality. However, they primarily target inference-time correction and do not systematically address how to organize correction differences as training-time process supervision---IOP fills precisely this gap.

\paragraph{Relationship to this work.}
The above lines respectively establish the effectiveness of outcome-supervised RL, exogenous process supervision, and self-correction. This paper addresses an orthogonal dimension: organizing the model's own correction capability into continuously improving internal process supervision during RL training, enabling fine-grained credit assignment without external step annotations or independent PRMs. IOP is complementary to any of the above lines---for instance, an exogenous PRM could initialize the repair reference or filter low-quality repairs.

\section{Method}
\label{sec:method}

\begin{figure}[h]
    \centering
    \resizebox{\textwidth}{!}{%
        \begin{tikzpicture}[
            >=Stealth,
            mainbox/.style={draw, rounded corners=4pt, minimum height=0.9cm, align=center, font=\footnotesize},
            policybox/.style={mainbox, fill=blue!8, draw=blue!50, minimum width=2.4cm},
            repairbox/.style={mainbox, fill=orange!10, draw=orange!60, minimum width=2.4cm},
            signalbox/.style={mainbox, fill=green!8, draw=green!50!black, minimum width=2.4cm},
            graybox/.style={mainbox, fill=gray!10, draw=gray!50, minimum width=2.2cm},
            updatebox/.style={mainbox, fill=violet!12, draw=violet!60, minimum width=3.6cm, minimum height=1.1cm, font=\footnotesize},
            synergybox/.style={draw, rounded corners=6pt, fill=yellow!12, draw=yellow!70!orange, thick, inner sep=5pt, align=center},
            tokenstyle/.style={minimum width=0.55cm, minimum height=0.55cm, font=\scriptsize, inner sep=1pt},
            tok_ok/.style={tokenstyle, fill=green!12, draw=green!35},
            tok_err/.style={tokenstyle, fill=red!18, draw=red!45},
            tok_fix/.style={tokenstyle, fill=orange!22, draw=orange!55},
            gate_on/.style={tokenstyle, fill=yellow!30, draw=yellow!65, font=\scriptsize\bfseries},
            gate_off/.style={tokenstyle, fill=gray!8, draw=gray!25, text=gray!45, font=\scriptsize},
            arr/.style={->, thick, >=Stealth},
            lblfont/.style={font=\scriptsize, text=black!65},
            secfont/.style={font=\footnotesize\bfseries},
            ]

            \node[secfont, text=red!65!black] (titleA) at (-6.5, 3.6)
            {(a) Existing: Outcome Supervision};

            \node[tok_ok]  (L1) at (-9.0, 2.9) {$z_1$};
            \node[tok_ok,  right=1pt of L1] (L2) {$z_2$};
            \node[tok_ok,  right=1pt of L2] (L3) {$z_3$};
            \node[tok_err, right=1pt of L3] (L4) {$z_4$};
            \node[tok_err, right=1pt of L4] (L5) {$z_5$};
            \node[tok_ok,  right=1pt of L5] (L6) {$z_6$};
            \node[tok_ok,  right=1pt of L6] (L7) {$z_7$};
            \node[tok_err, right=1pt of L7] (L8) {$z_8$};
            \node[right=4pt of L8, font=\scriptsize, text=red!65!black] {$r\!=\!0$};

            \draw[decorate, decoration={brace, amplitude=4pt, mirror}, thick, red!55]
            ([yshift=-2pt]L1.south west) -- ([yshift=-2pt]L8.south east)
            node[midway, below=5pt, font=\scriptsize, text=red!65!black, align=center]
            {Uniform negative gradient $\nabla^{-}$\enspace{\tiny cannot distinguish correct from incorrect tokens}};

            \draw[->, line width=2pt, violet!60, shorten >=4pt, shorten <=4pt]
            (-3.6, 2.3) -- (0.8, 2.3)
            node[midway, above, font=\scriptsize\bfseries, text=violet!70!black] {Internalize}
            node[midway, below, font=\tiny, text=violet!50!black] {sequence-level $\to$ token-level};

            \node[secfont, text=blue!65!black] (titleB) at (5.2, 3.6)
            {(b) New Perspective: Internalize Outcome into Process Supervision};

            \node[lblfont, text=red!55!black, anchor=east] at (2.35, 2.9) {Failed $y$};
            \node[tok_ok]  (R1) at (3.0, 2.9) {$y_1$};
            \node[tok_ok,  right=1pt of R1] (R2) {$y_2$};
            \node[tok_ok,  right=1pt of R2] (R3) {$y_3$};
            \node[tok_err, right=1pt of R3] (R4) {$y_4$};
            \node[tok_err, right=1pt of R4] (R5) {$y_5$};
            \node[tok_ok,  right=1pt of R5] (R6) {$y_6$};
            \node[tok_ok,  right=1pt of R6] (R7) {$y_7$};
            \node[tok_err, right=1pt of R7] (R8) {$y_8$};
            \node[right=4pt of R8, font=\scriptsize, text=red!65!black] {$r\!=\!0$};

            \node[lblfont, text=green!45!black, anchor=east] at (2.35, 2.15) {Repair $\tilde{y}$};
            \node[tok_ok]  (F1) at (3.0, 2.15) {${y}_1$};
            \node[tok_ok,  right=1pt of F1] (F2) {${y}_2$};
            \node[tok_ok,  right=1pt of F2] (F3) {${y}_3$};
            \node[tok_fix, right=1pt of F3] (F4) {$\tilde{y}_4$};
            \node[tok_fix, right=1pt of F4] (F5) {$\tilde{y}_5$};
            \node[tok_ok,  right=1pt of F5] (F6) {${y}_6$};
            \node[tok_ok,  right=1pt of F6] (F7) {${y}_7$};
            \node[tok_fix, right=1pt of F7] (F8) {$\tilde{y}_8$};
            \node[right=4pt of F8, font=\scriptsize, text=green!45!black] {$r\!=\!1$};

            \foreach \i in {4,5,8} {
                \draw[arr, orange!65, semithick] (R\i.south) -- (F\i.north);
            }
            \node[font=\tiny, text=orange!65!black, rotate=90] at ($(R4.south)!0.5!(F4.north)+(-0.32,0)$) {Align};

            \node[lblfont, anchor=east] at (2.35, 1.4) {Gate $g_t$};
            \node[gate_off] (G1) at (3.0, 1.4)  {0};
            \node[gate_off, right=1pt of G1] (G2) {0};
            \node[gate_off, right=1pt of G2] (G3) {0};
            \node[gate_on,  right=1pt of G3] (G4) {1};
            \node[gate_on,  right=1pt of G4] (G5) {1};
            \node[gate_off, right=1pt of G5] (G6) {0};
            \node[gate_off, right=1pt of G6] (G7) {0};
            \node[gate_on,  right=1pt of G7] (G8) {1};

            \draw[decorate, decoration={brace, amplitude=4pt, mirror}, thick, blue!50]
            ([yshift=-2pt]G1.south west) -- ([yshift=-2pt]G8.south east)
            node[midway, below=5pt, font=\scriptsize, text=blue!65!black, align=center]
            {Gradients only at diff positions\enspace{\tiny internalized token-level process supervision}};

            \node[secfont, text=black!70, align=center] at (-10.2, -1.0)
            {(c)\\New};

            \node[policybox, minimum height=0.9cm, minimum width=1.7cm] (pol) at (-8.5, -1.0)
            {Policy Mode\\[-1pt]{\scriptsize $\pi_\theta$}};
            \node[graybox, minimum height=0.9cm, minimum width=1.7cm] (spl) at (-6.0, -1.0)
            {Split Cor/Err\\[-1pt]{\scriptsize $\mathcal{G}_{\text{cor}}/\mathcal{G}_{\text{err}}$}};

            \node[repairbox, minimum height=1.4cm, minimum width=2.6cm, inner sep=3pt] (rep) at (-3.0, -1.0) {};
            \node[font=\scriptsize, anchor=north] at (rep.north)
            {Repair Mode $\rho_\theta$};
            \node[draw, rounded corners=2pt, fill=red!6, draw=red!40, font=\tiny,
            minimum height=0.35cm, minimum width=2.0cm, anchor=south] (aud)
            at ([yshift=2pt]rep.south)
            {Audit Gate $\mathcal{M}_{\text{audit}}$};

            \node[repairbox, minimum height=0.7cm, minimum width=1.7cm, font=\scriptsize] (brep)
            at (1.2, -0.5) {$\mathcal{B}_{\text{rep}}$};

            \node[signalbox, minimum height=0.7cm, minimum width=2.2cm, font=\scriptsize] (aln)
            at (0.2, -1.5) {Align+Trunc $g_t$};
            \node[policybox, minimum height=0.7cm, minimum width=1.7cm, font=\scriptsize] (bpol)
            at (3.2, -1.5) {$\mathcal{B}_{\text{pol}}$};

            \node[updatebox, minimum width=3.0cm, minimum height=1.4cm] (joint) at (6.5, -1.0)
            {Joint Update $\theta$\\[-1pt]{\scriptsize $\mathcal{J}_{\text{IOP}}\!=\!\mathcal{J}_{\text{GSPO}}\!+\!\lambda\mathcal{J}_{\text{rep}}$}};

            \draw[arr] (pol) -- (spl) node[midway, above, lblfont] {$G'$};
            \draw[arr] (spl) -- (rep) node[midway, above, lblfont] {$y{+}a$};
            \draw[arr, orange!60] (rep.east |- brep) -- (brep.west);
            \draw[arr] (rep.east |- aln) -- (aln.west)
            node[midway, above, lblfont] {$(y,\tilde{y})$};
            \draw[arr, blue!60] (aln) -- (bpol) node[midway, above, lblfont] {$g_t$};
            \draw[arr, orange!60, thick] (brep.east) -- (joint.west |- brep);
            \draw[arr, blue!60, thick] (bpol.east) -- (joint.west |- bpol);

            \node[font=\scriptsize, text=black!55] at (-2.0, -2.1)
            {\textbf{Synergistic reinforcement}: Policy$\uparrow$ $\rightarrow$ Reference quality$\uparrow$ $\rightarrow$ Repair precision$\uparrow$ $\rightarrow$ Gating focus$\uparrow$ $\rightarrow$ Policy$\uparrow$};

        \end{tikzpicture}
    }%
    \caption{Internalizing outcome supervision into process supervision. Existing paradigm $\to$ New perspective $\to$ New paradigm.}
    \label{fig:overview}
\end{figure}
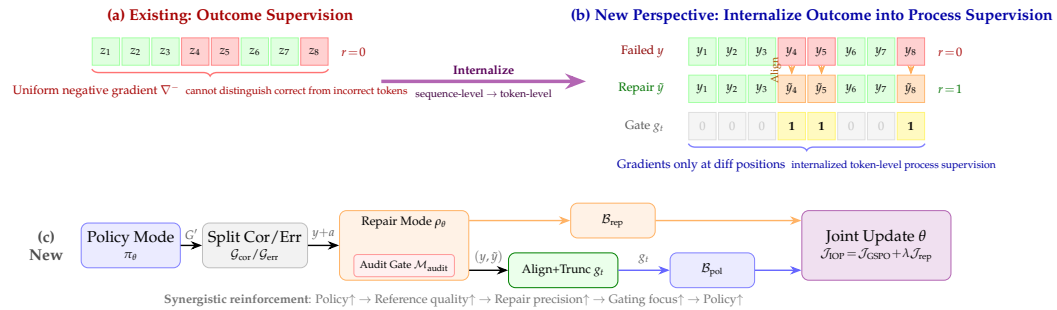

\paragraph{Overview.}
We refer to the proposed method as \textbf{IOP} (\emph{Internalizing Outcome Supervision into Process Supervision}), which comprises three core steps: (i) selecting repairable failed trajectories, (ii) generating internal process supervision through a repair mode, and (iii) feeding process signals back into policy updates via truncation gating. The gating mechanism is decoupled from the outer optimizer and can be embedded into any group-comparison algorithm such as GRPO, GSPO, or RLOO; we instantiate IOP with GSPO \citep{zheng2025groupsequencepolicyoptimization} as the outer optimizer, yielding \textbf{IOP-GSPO}.

A single model with shared parameters $\theta$ serves two roles: a \textbf{policy mode} $\pi_{\theta}(\mathbf{y}\mid x)$ that generates reasoning trajectories $\mathbf{y}=(y_1,\ldots,y_T)$ and receives binary scores $r(x,\mathbf{y})\!\in\!\{0,1\}$; and a \textbf{repair mode} $\rho_{\theta}(\tilde{\mathbf{y}}\mid x,\mathbf{y},a)$ that produces a repair $\tilde{\mathbf{y}}$ conditioned on a failed trajectory $\mathbf{y}$ and a reference $a$. Binary rewards lack discrimination at the sequence level, but the repair--alignment pipeline converts them into token-level difference signals. Parameter sharing avoids extra model overhead and ensures that repair capability improves in tandem with the policy: a stronger policy produces more concentrated errors, yielding more precise repairs and more focused gating.

\subsection{IOP-GSPO: Cold Start and Joint Optimization}
\label{sec:iop_grpo_impl}

IOP proceeds in two stages: Stage~1 endows the model with basic repair capability through cold-start SFT, and Stage~2 continuously converts failed trajectories into process supervision and feeds it back to the policy via joint RL.

\subsubsection{Stage 1: Acquiring Repair Capability via SFT}
\label{sec:repair_sft}

We first construct a cold-start repair dataset of 500 examples
\[
\mathcal{D}_{\text{rep}}^{0}=\{(x,\mathbf{y},a,\mathbf{y}^{*})\}
\]
and fine-tune the repair mode $\rho_{\theta}$ with SFT. The input is the concatenation of a repair instruction and $(x, \mathbf{y}, a)$ (prompt: ``Please correct the following erroneous reasoning with reference to the correct trajectory''), and the output is $\mathbf{y}^{*}$:
\begin{equation}
    \mathcal{L}_{\text{SFT}}(\theta)
    =
    - \mathbb{E}_{(x,\mathbf{y},a,\mathbf{y}^{*})\sim \mathcal{D}_{\text{rep}}^{0}}
    \Bigg[
    \sum_{t=1}^{T^*}
    \log \rho_{\theta}(y_t^{*}\mid x,\mathbf{y},a,\mathbf{y}^{*}_{<t})
    \Bigg].
\end{equation}
The dataset is constructed by sampling from the base model, filtering incorrect--correct pairs, and manually writing repair trajectories. 500 examples suffice to learn the local correction format; subsequent RL continuously improves repair quality and coverage (38.4\%$\to$73.1\%, Table~\ref{tab:dynamics}).

\subsubsection{Stage 2: Joint RL Optimization of Policy and Repair}
\label{sec:repair_rl}

Each step executes: sampling $\to$ correct/incorrect partition $\to$ repair $\to$ dual-batch construction $\to$ joint update. Failed trajectories are simultaneously written to $\mathcal{B}_{\text{rep}}$ (for repair capability) and $\mathcal{B}_{\text{pol}}$ (for localized policy updates), so that both capabilities are jointly updated at every step.

\paragraph{Sampling the initial group.}
For each prompt $x$, we sample $G'$ candidate trajectories from the old policy $\pi_{\theta_{\text{old}}}$:

\begin{equation}
    \mathcal{G}'(x)=\{y_i\}_{i=1}^{G'},\qquad
    y_i \sim \pi_{\theta_{\text{old}}}(\cdot\mid x),
    \label{eq:Gprime_sampling}
\end{equation}

\paragraph{Correct/incorrect partition.}
Trajectories are partitioned by reward threshold $\tau_r$ (with binary rewards, $\tau_r\!=\!0.5$):
\begin{equation}
    \mathcal{G}_{\text{cor}}(x)=\{y\in \mathcal{G}'(x)\mid r(x,y)\ge \tau_r\},\qquad
    \mathcal{G}_{\text{err}}(x)=\{y\in \mathcal{G}'(x)\mid r(x,y)< \tau_r\}.
    \label{eq:split_Gprime}
\end{equation}

\paragraph{Reference trajectory.}
If the error set is empty, the prompt is skipped. If the correct set is empty, the prompt is deferred until the model's capability improves.
When both sets are non-empty, a reference is sampled uniformly $a \sim \mathrm{Unif}(\mathcal{G}_{\text{cor}}(x))$ to avoid bias toward any particular solution.

\paragraph{Design choice: correct samples serve only as reference anchors.}
$\mathcal{G}_{\text{cor}}(x)$ provides only the reference anchor $a$ and does not participate in policy gradient computation. Including correct samples concentrates probability mass on a few trajectories and induces entropy collapse (HMMT25 drops 6.1\%, Table~\ref{tab:ablation}).

\paragraph{Repairing failed trajectories.}
For each $y_i\!\in\!\mathcal{G}_{\text{err}}(x)$, $G_{\text{rep}}$ candidate repairs are generated using $a$ as the anchor:
\begin{equation}
    \widetilde{\mathcal{G}}_{\text{rep}}(x,y_i,a)
    =
    \{\tilde y_i^{(j)}\}_{j=1}^{G_{\text{rep}}},
    \qquad
    \tilde y_i^{(j)} \sim \rho_{\theta}(\cdot \mid x, y_i, a),
    \label{eq:repair_each_error}
\end{equation}
where $r_{\text{task}}(x,\tilde y)\!\in\!\{0,1\}$ denotes the correctness judgment.

\paragraph{Reward hacking audit gate.}
Each candidate repair is subject to an external audit gate $h_i^{(j)}\!\in\!\{0,1\}$: the original failed trajectory $y_i$, candidate repair $\tilde y_i^{(j)}$, and reference trajectory $a$ are submitted to an independent audit model $\mathcal{M}_{\text{audit}}$ (we use OpenAI's open-source gpt-oss-120b \citep{openai2025gptoss120bgptoss20bmodel}), avoiding circular reasoning from parameter sharing:
\begin{equation}
    h_i^{(j)}
    =
    \mathcal{M}_{\text{audit}}(x, y_i, \tilde y_i^{(j)}, a)
    \in \{0,1\},
    \label{eq:rh_audit}
\end{equation}
where $h_i^{(j)}\!=\!1$ indicates passing the audit (no reward hacking) and $h_i^{(j)}\!=\!0$ indicates detected reward hacking.

Candidates are jointly scored by the audit gate, correctness, and normalized edit distance $\bar{\Delta}_{\text{edit}}$, favoring repairs that pass the audit, are correct, and involve minimal edits:
\begin{equation}
    s_{\text{rep}}(x,y_i,a,\tilde y_i^{(j)})
    =
    h_i^{(j)}
    \cdot
    \Big(
    r_{\text{task}}(x,\tilde y_i^{(j)})
    -
    \lambda_{\text{edit}}\,
    \bar{\Delta}_{\text{edit}}(y_i,\tilde y_i^{(j)})
    \Big).
    \label{eq:repair_score}
\end{equation}
\paragraph{Selecting the best repair.}
The highest-scoring candidate is selected (ties broken in favor of $r_{\text{task}}\!=\!1$):
\begin{equation}
    \tilde y_i
    =
    \arg\max_{\tilde y_i^{(j)}\in \widetilde{\mathcal{G}}_{\text{rep}}(x,y_i,a)}
    s_{\text{rep}}(x,y_i,a,\tilde y_i^{(j)}).
    \label{eq:rs_optimal}
\end{equation}
If the best repair has $r_{\text{task}}\!=\!0$, the prompt is deferred; otherwise, the repair candidate group is written to $\mathcal{B}_{\text{rep}}$, and the best repair paired with the original error forms a two-element tuple written to $\mathcal{B}_{\text{pol}}$. The pairwise structure ($|\mathcal{G}_i|\!=\!2$) maximizes the discriminability of difference masks: token-level comparison directly localizes the key positions that change the answer:
\begin{equation}
    \mathcal{G}_i(x)=\{y_i,\tilde y_i\},
    \quad
    \mathcal{B}_{\text{pol}}(x)
    =
    \{\mathcal{G}_i(x)\mid y_i\in\mathcal{G}_{\text{err}}(x)\},
    \quad
    |\mathcal{G}_i(x)| = 2.
    \label{eq:pair_group}
\end{equation}
The corresponding repair mini-batch is
\[
\mathcal{B}_{\text{rep}}(x)=\{(x,y_i,a,\widetilde{\mathcal{G}}_{\text{rep}}(x,y_i,a))\mid y_i\in\mathcal{G}_{\text{err}}(x)\}.
\]

\subsubsection{Process Supervision Signal Generation: Alignment and Truncation}
\label{sec:align_mask}

The alignment operator $\mathcal{A}$ computes bilateral difference masks $(\mathbf{m},\mathbf{m}')=\mathcal{A}(\mathbf{y},\tilde{\mathbf{y}})$ offline along the token-level Levenshtein path, where $\mathbf{m}\!\in\!\{0,1\}^{T}$ and $\mathbf{m}'\!\in\!\{0,1\}^{\tilde T}$ mark substitution, deletion, or insertion positions, respectively.

\paragraph{Truncated difference mask.}
Early errors induce cascade drift---a single local error triggers global token-level divergence downstream---and full-sequence updates propagate negative feedback to unrelated positions. To mitigate this, we retain only the earliest $K$ difference positions and construct a \textbf{truncated mask}:
\begin{equation}
    m_t^{(K)}=
    \begin{cases}
        1, & t\in \mathcal{S}_{K}(\mathbf{y},\tilde{\mathbf{y}}),\\
        0, & \text{otherwise},
    \end{cases}
    \label{eq:mask_trunc}
\end{equation}
with $m_t^{\prime(K)}$ defined analogously for the repair side. $\mathcal{S}_{K}$ denotes the first $K$ difference tokens selected in positional order.

\paragraph{Verification-based adaptive truncation.}
A fixed $K$ may be insufficient to complete the repair. We therefore construct a \textbf{grafted trajectory}: the tokens at positions $\mathcal{S}_K$ in $\mathbf{y}$ are replaced with the corresponding tokens from $\tilde{\mathbf{y}}$, and generation continues from $t_K^{\max}\!=\!\max(\mathcal{S}_K)$:
\begin{equation}
    \hat{y}_t^{(K)}=
    \begin{cases}
        \tilde{y}_t, & t \in \mathcal{S}_K,\\
        y_t, & t \le t_K^{\max},\; t \notin \mathcal{S}_K,\\
        \sim \pi_\theta(\cdot \mid x, \hat{y}_{<t}^{(K)}), & t > t_K^{\max}.
    \end{cases}
    \label{eq:graft}
\end{equation}
If $r_{\text{task}}(x,\hat{\mathbf{y}}^{(K)})\!=\!1$, the $K$ edits are verified as sufficient; otherwise the window is expanded to $2K$; if this also fails, the full mask is used:
\begin{equation}
    K^{*} =
    \begin{cases}
        K, & r_{\text{task}}(x, \hat{\mathbf{y}}^{(K)}) = 1,\\
        2K, & r_{\text{task}}(x, \hat{\mathbf{y}}^{(K)}) = 0 \;\wedge\; r_{\text{task}}(x, \hat{\mathbf{y}}^{(2K)}) = 1,\\
        |\mathcal{S}_{\infty}|, & \text{otherwise}.
    \end{cases}
    \label{eq:adaptive_trunc}
\end{equation}
Empirically, approximately 72\% of samples pass at $K$, 19\% fall back to $2K$, and 9\% use the full mask (Table~\ref{tab:ablation}).

The resulting token-level gating is:
\begin{equation}
    g_t \in [0,1],\qquad
    g_t = m_t^{(K^*)} \ \ \text{or}\ \ g_t=m_t^{\prime(K^*)},
    \label{eq:gating}
\end{equation}

\subsection{Gated Policy Optimization and Joint Training}
\label{sec:df_gspo}

For each $\mathcal{G}_i(x)$ in $\mathcal{B}_{\text{pol}}(x)$, we write
\[
\mathcal{G}_i(x)=\{y_i,\tilde y_i\},
\]
and let $z\in\mathcal{G}_i(x)$ denote either trajectory in the pair, with the sequence-level advantage $\widehat{A}_i(z)$ computed via within-pair normalization.

\paragraph{Bilateral difference gating.}
For the pair $\{y_i,\tilde y_i\}$, the truncated alignment operator $\mathcal{A}_K$ simultaneously produces error-side and repair-side gating:
\begin{equation}
    \big(m_{i,1:|y_i|}^{(K)},\ m_{i,1:|\tilde{y}_i|}^{\prime(K)}\big)
    =
    \mathcal{A}_{K}(y_i,\tilde{y}_i),
    \label{eq:mask_pair_two_sided}
\end{equation}
with the corresponding token-level gating
\begin{equation}
    g_{z,t}(x)=
    \begin{cases}
        m_{i,t}^{(K)},  & z = y_i,\\[3pt]
        m_{i,t}^{\prime(K)}, & z = \tilde{y}_i.
    \end{cases}
    \label{eq:pair_two_sided_gating}
\end{equation}

GSPO uses a sequence-level likelihood ratio; IOP introduces gating on top, restricting the sequence-level ratio to only those tokens at difference positions:
\begin{equation}
    w_{z}^{\text{gate}}(\theta)
    =
    \exp\!\Bigg(\frac{1}{\sum_{t'=1}^{T_z} g_{z,t'}}\sum_{t=1}^{T_z} g_{z,t}\,\log\frac{\pi_{\theta}(z_t\mid x, z_{<t})}
    {\pi_{\theta_{\text{old}}}(z_t\mid x, z_{<t})}\Bigg).
    \label{eq:ratio_z_gate}
\end{equation}

Following the token-level importance ratio of GSPO-token \citep{zheng2025groupsequencepolicyoptimization}:
\begin{equation}
    s_{z,t}(\theta)
    =
    \mathrm{sg}\!\big[w_{z}^{\text{gate}}(\theta)\big]
    \cdot
    \frac{\pi_{\theta}(z_t\mid x, z_{<t})}
    {\mathrm{sg}\!\big[\pi_{\theta}(z_t\mid x, z_{<t})\big]},
    \label{eq:token_ratio_gate}
\end{equation}
where $\mathrm{sg}[\cdot]$ denotes the stop-gradient operator that takes the numerical value while blocking gradient flow. $s_{z,t}$ is numerically equal to $w_{z}^{\text{gate}}$, but the gradient propagates only through the log-probability at position $t$. The gated token-level clipped objective is averaged over active positions:
\begin{equation}
    \mathcal{L}_{z}^{\text{clip}}
    =
    \frac{1}{\sum_{t'=1}^{T_z} g_{z,t'}}
    \sum_{t=1}^{T_z} g_{z,t}\,
    \min\Big(
    s_{z,t}(\theta)\,\widehat{A}_i(z),\;
    \mathrm{clip}\big(s_{z,t}(\theta),1-\epsilon,1+\epsilon\big)\,\widehat{A}_i(z)
    \Big).
    \label{eq:clip_loss_pair}
\end{equation}
Dividing by $\sum_{t'} g_{z,t'}$ (the actual number of active tokens) ensures that the gradient scale is invariant to mask sparsity, serving the same role as the $1/|y_i|$ normalization in GSPO. The IOP-GSPO policy objective is
\begin{align}
    \mathcal{J}_{\text{IOP-GSPO}}(\theta;\mathcal{B}_{\text{pol}})
    &=
    \mathbb{E}_{x \sim \mathcal{D},\, \mathcal{G}_i(x)\sim \mathcal{B}_{\text{pol}}(x)}
    \Bigg[
    \frac{1}{2}
    \sum_{z\in\mathcal{G}_i(x)}
    \mathcal{L}_{z}^{\text{clip}}
    \Bigg]
    -
    \beta_{\text{kl}}\,\mathcal{K}(\theta),
    \label{eq:df_gspo}
\end{align}
\paragraph{Repair mini-batch objective.}
For each $(x,y_i,a,\widetilde{\mathcal{G}}_{\text{rep}})$ in $\mathcal{B}_{\text{rep}}(x)$, the candidate group
\[
\widetilde{\mathcal{G}}_{\text{rep}}(x,y_i,a)=\{\tilde y_i^{(j)}\}_{j=1}^{G_{\text{rep}}}
\]
is optimized in GSPO form. Let $\widehat{A}^{\text{rep}}_i(\tilde y_i^{(j)})$ denote the within-group advantage normalized by $s_{\text{rep}}$. The repair sequence-level likelihood ratio is
\begin{equation}
    u_i^{(j)}(\theta)
    =
    \exp\!\Bigg(\frac{1}{T_{\tilde y_i^{(j)}}}\sum_{t=1}^{T_{\tilde y_i^{(j)}}}
    \log\frac{\rho_{\theta}(\tilde y_{i,t}^{(j)}\mid x, y_i, a, \tilde y_{i,<t}^{(j)})}
    {\rho_{\theta_{\text{old}}}(\tilde y_{i,t}^{(j)}\mid x, y_i, a, \tilde y_{i,<t}^{(j)})}\Bigg).
    \label{eq:repair_ratio}
\end{equation}
The repair sequence-level clipped term is
\begin{equation}
    \mathcal{L}_{i}^{\text{rep},(j)}
    =
    \min\Big(
    u_i^{(j)}(\theta)\,\widehat{A}^{\text{rep}}_i(\tilde y_i^{(j)}),\;
    \mathrm{clip}\big(u_i^{(j)}(\theta),1-\epsilon,1+\epsilon\big)\,\widehat{A}^{\text{rep}}_i(\tilde y_i^{(j)})
    \Big).
    \label{eq:repair_clip}
\end{equation}
The repair objective optimizes over $G_{\text{rep}}$ candidates in GSPO form, steering the repair mode toward high-scoring candidates:
\begin{align}
    \mathcal{J}_{\text{rep}}(\theta;\mathcal{B}_{\text{rep}})
    &=
    \mathbb{E}_{(x,y_i,a,\widetilde{\mathcal{G}}_{\text{rep}})\sim \mathcal{B}_{\text{rep}}}
    \Bigg[
    \frac{1}{G_{\text{rep}}}
    \sum_{j=1}^{G_{\text{rep}}}
    \mathcal{L}_{i}^{\text{rep},(j)}
    \Bigg]
    -
    \beta_{\text{kl}}\,\mathcal{K}(\theta),
    \label{eq:repair_gspo}
\end{align}

Each step jointly updates the shared parameters $\theta$:
\begin{equation}
    \mathcal{J}_{\text{IOP}}(\theta;\mathcal{B}_{\text{pol}},\mathcal{B}_{\text{rep}})
    =
    \mathcal{J}_{\text{IOP-GSPO}}(\theta;\mathcal{B}_{\text{pol}})
    +
    \lambda_{\text{rep}}\,\mathcal{J}_{\text{rep}}(\theta;\mathcal{B}_{\text{rep}}),
\end{equation}
where $\lambda_{\text{rep}}$ controls the repair weight (sensitivity analysis in Appendix~\ref{app:lambda_sensitivity}) and $\mathcal{K}(\theta)$ is the token-level KL regularizer against the reference policy. The complete single-step procedure is given in Appendix Algorithm~1.

\section{Experiments}
\label{sec:experiments}

\subsection{Experimental Setup}
\label{subsec:setup}

We compare IOP-GSPO with GSPO under a compute-matched setting (identical total token budget). The evaluation metric is Acc avg@32 (average accuracy over 32 independent samples). Base models are Qwen3-32B (dense) and Qwen3-Next-80B-A3B-Thinking (MoE, 3B active parameters), covering both dense and MoE architectures. Training data is a mixed subset of DeepMath-103K \citep{he2026deepmathk} (decontaminated mathematical reasoning dataset) and OpenCodeReasoning \citep{ahmad2025opencodereasoning} (decontaminated code reasoning dataset); cold-start data is constructed as described in \S\ref{sec:repair_sft}. Evaluation benchmarks span three reasoning categories: AIME25~\cite{maa_aime2025} (competition mathematics), HMMT25~\cite{balunovic2026matharena} (long-chain mathematical reasoning), and LiveCodeBench v6~\cite{jain2025livecodebench} (code reasoning). Full hyperparameters are provided in Appendix Table~\ref{tab:hyperparams}.

\subsection{Main Results}
\label{subsec:main_results}

\begin{table}[t]
    \centering
    \small
    \setlength{\tabcolsep}{6pt}
    \renewcommand{\arraystretch}{1.15}
    \resizebox{\textwidth}{!}{
    \begin{tabular}{lcccccc}
        \toprule
        \multirow{2}{*}{\textbf{Method}} &
        \multicolumn{3}{c|}{\textbf{Qwen3-32B Acc avg@32 (\%)}} &
        \multicolumn{3}{c}{\textbf{Qwen3-Next Acc avg@32 (\%)}} \\
        \cmidrule(lr){2-4}\cmidrule(lr){5-7}
        & \textbf{AIME25} & \textbf{LiveCodeBench} & \textbf{HMMT25}
        & \textbf{AIME25} & \textbf{LiveCodeBench} & \textbf{HMMT25} \\
        \midrule
        Base      & 72.9  & 60.6 & 51.5 & 87.8 & 68.7 & 73.9 \\
        GSPO      & 76.6$\pm$1.1 & 61.8$\pm$0.9 & 55.4$\pm$1.2 & 88.4$\pm$1.4 & 69.6$\pm$0.8 & 75.4$\pm$1.4 \\
        IOP-GSPO   & \textbf{83.5$\pm$1.1} & \textbf{67.8$\pm$1.1} & \textbf{63.2$\pm$1.1} & \textbf{92.1$\pm$1.2} & \textbf{73.8$\pm$1.5} & \textbf{82.2$\pm$0.8} \\
        \bottomrule
    \end{tabular}
    }
    \caption{Main results under compute-matched setting (mean $\pm$ 95\% bootstrap CI over 5 seeds). IOP-GSPO consistently outperforms GSPO across all three benchmarks and both architectures (dense and MoE); all confidence intervals are non-overlapping, indicating statistically significant differences.}
    \label{tab:main_llm_results}
\end{table}

Table~\ref{tab:main_llm_results} shows that IOP-GSPO consistently outperforms GSPO across all configurations (Qwen3-32B average +6.9\%, Qwen3-Next average +4.9\%). Improvements are most pronounced on long-chain reasoning (HMMT25: +7.8\%/+6.8\%), where localized gating has the greatest advantage. Additionally, under a step-matched comparison, IOP-GSPO reaches the final performance of GSPO at 800 steps in only approximately 350 steps, yielding approximately $2.3\times$ sample efficiency (Appendix~\ref{app:compute_cost}).

\subsection{Ablation Analysis and Method Comparison}
\label{subsec:analysis}

\begin{table}[t]
    \centering
    \small
    \setlength{\tabcolsep}{5pt}
    \renewcommand{\arraystretch}{1.15}
    \begin{tabular}{lcc}
        \toprule
        \textbf{Method / Variant} & \textbf{AIME25} & \textbf{HMMT25} \\
        \midrule
        IOP-GSPO (full) & \textbf{83.5} & \textbf{63.2} \\
        \midrule
        \multicolumn{3}{l}{\emph{Ablation variants (removing one component at a time)}} \\
        \ \ $-$ Fixed truncation (no adaptive verification) & 82.1 & 61.8 \\
        \ \ $-$ No truncation ($K\!=\!\infty$) & 79.8 & 58.6 \\
        \ \ $-$ No minimum-edit constraint & 80.1 & 59.4 \\
        \ \ $-$ Correct samples in gradient updates & 78.3 & 57.1 \\
        \ \ $-$ No gating (full-token updates) & 80.6 & 59.9 \\
        \ \ $-$ No audit gate ($h\!\equiv\!1$) & 81.2 & 60.5 \\
        \midrule
        \multicolumn{3}{l}{\emph{Exogenous process supervision methods}} \\
        GSPO + PRM rerank (step-level annotations) & 79.4 & 58.8 \\
        PRIME \citep{cui2025processreinforcementimplicitrewards}  & 78.5 & 58.1 \\
        V-PPO \citep{liu2026savegoodprefixprecise} (verified prefix) & 80.2 & 59.6 \\
        \midrule
        \multicolumn{3}{l}{\emph{Sampling baselines}} \\
        GSPO ($G'\!=\!16$) & 76.6 & 55.4 \\
        GSPO ($G'\!=\!48$, compute-matched) & 77.8 & 56.9 \\
        GSPO + rejection sampling & 78.1 & 57.3 \\
        \bottomrule
    \end{tabular}
    \caption{Ablation and method comparison on Qwen3-32B (Acc avg@32, \%). Top: all six design ablations cause significant degradation; middle: IOP-GSPO outperforms PRM rerank, PRIME, and V-PPO without external annotations; bottom: increased sampling alone does not explain the gains.}
    \label{tab:ablation}
\end{table}

Table~\ref{tab:ablation} validates the necessity of each of the six design components on Qwen3-32B. The most impactful is \textbf{correct sample exclusion}: including correct samples causes probability mass to concentrate too rapidly, inducing entropy collapse mid-training (HMMT25 drops 6.1\%). Removing the \textbf{minimum-edit constraint} degrades repair into wholesale rewriting; disabling \textbf{truncation} allows cascade drift to propagate negative feedback to unrelated positions; removing \textbf{gating} reduces the method to sequence-level advantage weighting; removing the \textbf{audit gate} ($h\!\equiv\!1$) allows approximately 8\% of reward-hacking repairs into training, introducing spurious process supervision signals (AIME25 drops 2.3\%). Replacing \textbf{verification-based adaptive truncation} with fixed-$K$ truncation (without graft verification) causes a 1.4\% drop on AIME25---approximately 28\% of samples actually require a larger truncation window, and fixed $K$ introduces insufficient training signals for these cases.

The middle section shows that IOP-GSPO outperforms GSPO + PRM rerank, PRIME, and V-PPO without any external annotations. The bottom section rules out the ``more sampling is sufficient'' hypothesis: compute-matched $G'\!=\!48$ and rejection sampling both fall well short of IOP-GSPO, confirming that the gains stem from signal quality rather than sampling volume.

\paragraph{Training dynamics.}
Appendix Table~\ref{tab:dynamics} shows that repair success rate rises from 38.4\% to 73.1\%, policy accuracy from 52.1\% to 83.5\%, and active token ratio decreases from 24.6\% to 9.8\%. The three metrics evolve in concert: improved policy produces better references that in turn improve repair quality, and more precise repairs improve local update signals. The steadily declining active token ratio indicates that gating automatically refines over training---direct evidence that outcome supervision is being internalized into process supervision. Hyperparameter sensitivity and computational overhead are discussed in Appendix~\ref{app:lambda_sensitivity}--\ref{app:compute_cost}.

\section{Conclusion}
\label{sec:conclusion}

We propose a new paradigm---internalizing outcome supervision into process supervision---and realize it through the IOP framework and its instantiation IOP-GSPO. The core idea is to have the same model generate minimum-edit repairs for failed trajectories, then convert outcome-level feedback into token-level gating signals via truncated alignment and audit gating. Experiments show that IOP-GSPO consistently outperforms GSPO (+4.9--6.9\%) and exogenous process supervision methods across three reasoning benchmarks, with policy and repair capabilities forming a synergistic reinforcement loop. IOP demonstrates a self-improvement pathway: supervision quality improves in tandem with policy capability, providing an effective and low-cost alternative for fine-grained credit assignment under outcome feedback alone. Future directions include extension to multi-turn agent tasks.

\section{Limitations}
\label{sec:limitations}

IOP has the following limitations. (1) Minimum edit distance does not equate to causal minimality---paraphrasing or equivalent transformations may cause the difference mask to deviate from the true error source; the method should therefore be viewed as a practical approximation rather than strict causal localization. (2) The method requires at least one correct trajectory per prompt as a reference anchor, and the repair mode must reliably produce high-quality candidates; both conditions are difficult to satisfy when the base model is too weak. (3) Experiments are limited to math and code reasoning benchmarks (up to 32B dense parameters) and do not cover larger-scale models, long-context reasoning, tool use, or multi-turn agent scenarios. 


\section*{Reproducibility Statement}
Code will be released.

\appendix
\section{Appendix}

\subsection{Algorithm}

\begin{algorithm}[htbp]
    \caption{IOP-GSPO Stage 2 single-step procedure: sampling $\to$ partition $\to$ repair $\to$ gating $\to$ joint update}
    \begin{algorithmic}[1]
        \State \textbf{Input:} prompt set $\mathcal{D}$, shared-parameter model $\theta$ (with policy mode and repair mode)
        \For{each prompt $x \in \mathcal{D}$}
            \State Sample initial group $\mathcal{G}'(x)=\{y_i\}_{i=1}^{G'}$, where $y_i \sim \pi_{\theta}(\cdot \mid x)$
            \State Partition $\mathcal{G}'(x)$ into $\mathcal{G}_{\text{cor}}(x)$ and $\mathcal{G}_{\text{err}}(x)$ by reward threshold
            \If{$\mathcal{G}_{\text{cor}}(x)=\emptyset$ or $\mathcal{G}_{\text{err}}(x)=\emptyset$}
                \State Skip the IOP pipeline for this prompt
            \Else
                \State Sample reference trajectory $a$ from $\mathcal{G}_{\text{cor}}(x)$
                \For{each $y_i \in \mathcal{G}_{\text{err}}(x)$}
                    \State Generate repair candidate group $\widetilde{\mathcal{G}}_{\text{rep}}(x,y_i,a)=\{\tilde{\mathbf{y}}^{(j)}\}_{j=1}^{G_{\text{rep}}}$
                    \State Score by $s_{\text{rep}}$ and select best repair $\tilde y_i$
                    \State Write $(x,y_i,a,\widetilde{\mathcal{G}}_{\text{rep}}(x,y_i,a))$ to repair mini-batch $\mathcal{B}_{\text{rep}}$
                    \State \textbf{If} $r_{\text{task}}(x,\tilde y_i)\!=\!1$: align and truncate to obtain mask, construct pair $\mathcal{G}_i(x)\!=\!\{y_i,\tilde y_i\}$
                \EndFor
                \State Write all pairs to policy mini-batch $\mathcal{B}_{\text{pol}}(x)=\{\mathcal{G}_i(x)\}$
                \State Compute IOP-GSPO policy objective from $\mathcal{B}_{\text{pol}}(x)$
                \State Compute standard GSPO objective for repair mode from $\mathcal{B}_{\text{rep}}(x)$
                \State Jointly update shared parameters $\theta$
            \EndIf
        \EndFor
    \end{algorithmic}
\end{algorithm}

\subsection{Training Dynamics}
\label{app:dynamics}

\begin{table}[htbp]
    \centering
    \small
    \setlength{\tabcolsep}{5pt}
    \renewcommand{\arraystretch}{1.15}
    \begin{tabular}{lccc}
        \toprule
        \textbf{Training Stage} & \textbf{Repair Success Rate (\%)} & \textbf{Policy Accuracy (\%)} & \textbf{Active Token Ratio (\%)} \\
        \midrule
        Step 0 (post-SFT) & 38.4 & 52.1 & 24.6 \\
        Step 200 & 51.7 & 69.0 & 15.8 \\
        Step 400 & 62.3 & 79.5 & 12.2 \\
        Step 600 & 68.9 & 81.8 & 10.6 \\
        Step 800 & 73.1 & 83.5 & 9.8 \\
        \bottomrule
    \end{tabular}
    \caption{Qwen3-32B training dynamics: synergistic reinforcement between policy and repair. Repair success rate (+34.7\%) and policy accuracy (+31.4\%) rise in tandem, while active token ratio steadily decreases (24.6\%$\to$9.8\%), indicating that gating signals automatically refine over training.}
    \label{tab:dynamics}
\end{table}

The post-cold-start repair success rate of only 38.4\% confirms the necessity of Stage~2 joint RL. Repair success rate improves most rapidly during Steps 0--200 (+13.3\%), coinciding with the period of fastest policy accuracy improvement, directly corroborating synergistic reinforcement. The slowdown in later stages reflects natural convergence: as policy accuracy improves, the number of available failed trajectories decreases, correspondingly reducing repair opportunities.

\subsection{Training Hyperparameters}
\label{app:hyperparams}

\begin{table}[htbp]
    \centering
    \small
    \setlength{\tabcolsep}{5pt}
    \renewcommand{\arraystretch}{1.15}
    \begin{tabular}{lcc}
        \toprule
        \textbf{Hyperparameter} & \textbf{Qwen3-32B} & \textbf{Qwen3-Next} \\
        \midrule
        Initial sampling group size $G'$ & 16 & 16 \\
        Repair candidates $G_{\text{rep}}$ & 4 & 4 \\
        Truncation length $K$ & 50 & 50 \\
        Edit distance weight $\lambda_{\text{edit}}$ (min.\ threshold 0.05) & 0.3 & 0.3 \\
        Repair objective weight $\lambda_{\text{rep}}$ & 0.2 & 0.2 \\
        KL penalty $\beta_{\text{kl}}$ & 0.002 & 0.002 \\
        Reward threshold $\tau_r$ & 0.5 & 0.5 \\
        Learning rate (AdamW, 20-step warmup) & $1\times10^{-6}$ & $5\times10^{-7}$ \\
        Prompt batch size & 64 & 64 \\
        Cold-start SFT data size & 500 & 500 \\
        Maximum sequence length & 32768 & 32768 \\
        \bottomrule
    \end{tabular}
    \caption{Complete IOP-GSPO training hyperparameters. The two architectures differ only in learning rate, indicating robustness to model scale and architecture type.}
    \label{tab:hyperparams}
\end{table}

\subsection[Sensitivity of Repair Objective Weight]{Sensitivity of Repair Objective Weight $\lambda_{\text{rep}}$}
\label{app:lambda_sensitivity}

\begin{table}[htbp]
    \centering
    \small
    \setlength{\tabcolsep}{6pt}
    \renewcommand{\arraystretch}{1.15}
    \begin{tabular}{lcc}
        \toprule
        $\lambda_{\text{rep}}$ & \textbf{AIME25} & \textbf{HMMT25} \\
        \midrule
        0.1 & 81.4 & 60.3 \\
        0.15 & 82.8 & 62.1 \\
        0.2 & \textbf{83.5} & \textbf{63.2} \\
        0.25 & 82.1 & 61.4 \\
        0.3 & 79.6 & 58.2 \\
        \bottomrule
    \end{tabular}
    \caption{Sensitivity of repair objective weight $\lambda_{\text{rep}}$ (Qwen3-32B, Acc avg@32, \%). All values in $\lambda_{\text{rep}}\!\in\![0.1,0.3]$ outperform the GSPO baseline.}
    \label{tab:lambda_sensitivity}
\end{table}

When $\lambda_{\text{rep}}$ is too small, repair training is insufficient (at 0.1, AIME25 reaches only 81.4\%); when too large, repair gradients dominate the shared parameters and degrade policy performance. $\lambda_{\text{rep}}\!=\!0.2$ achieves the best balance between policy and repair.

\subsection[Sensitivity of Truncation Length K]{Sensitivity of Truncation Length $K$}
\label{app:k_sensitivity}

\begin{table}[htbp]
    \centering
    \small
    \setlength{\tabcolsep}{6pt}
    \renewcommand{\arraystretch}{1.15}
    \begin{tabular}{lccc}
        \toprule
        \textbf{Truncation length $K$} & \textbf{AIME25} & \textbf{HMMT25} & \textbf{LiveCodeBench} \\
        \midrule
        $K=10$  & 81.2 & 60.8 & 65.2 \\
        $K=50$  & \textbf{83.5} & \textbf{63.2} & \textbf{67.8} \\
        $K=80$ & 82.7 & 62.1 & 67.1 \\
        $K=160$ & 81.0 & 59.7 & 65.6 \\
        $K=\infty$ & 79.8 & 58.6 & 64.2 \\
        \bottomrule
    \end{tabular}
    \caption{Effect of truncation length $K$ on IOP-GSPO performance (Qwen3-32B, Acc avg@32, \%). 
    }
    \label{tab:k_sensitivity}
\end{table}


\subsection[Effect of Repair Candidate Count]{Effect of Repair Candidate Count $G_{\text{rep}}$}
\label{app:grep_sensitivity}

\begin{table}[htbp]
    \centering
    \small
    \setlength{\tabcolsep}{6pt}
    \renewcommand{\arraystretch}{1.15}
    \begin{tabular}{lccc}
        \toprule
        \textbf{$G_{\text{rep}}$} & \textbf{AIME25} & \textbf{HMMT25} & \textbf{Repair Success Rate (\%)} \\
        \midrule
        2 & 82.1 & 61.5 & 59.6 \\
        4 & \textbf{83.5} & \textbf{63.2} & 73.1 \\
        8 & 83.2 & 63.0 & 82.4 \\
        \bottomrule
    \end{tabular}
    \caption{Effect of repair candidate count $G_{\text{rep}}$ on task performance and repair success rate (Qwen3-32B, Acc avg@32, \%).}
    \label{tab:grep_sensitivity}
\end{table}

Increasing $G_{\text{rep}}$ from 2 to 4 raises repair success rate from 59.6\% to 73.1\%, with task performance improving correspondingly; further increase to 8 yields diminishing returns while linearly increasing cost. $G_{\text{rep}}\!=\!4$ provides the best balance between performance and efficiency.

\subsection{Computational Overhead}
\label{app:compute_cost}

The additional overhead of IOP-GSPO primarily comes from repair sampling. As training progresses, the error rate decreases from approximately 48\% initially to approximately 17\% later (Table~\ref{tab:dynamics}); the training-weighted average error rate is approximately 26\%, generating approximately 17 additional repair candidates per input ($\approx$4.2 failed trajectories $\times\,G_{\text{rep}}\!=\!4$), resulting in approximately $2.2\times$ token throughput increase after accounting for audit gating and graft verification. The GSPO baseline's training steps are proportionally increased to consume the same total token budget (compute-matched setting). Under step-matched comparison, IOP-GSPO reaches GSPO's final 800-step performance in only approximately 350 steps (AIME25: 76.8\% vs.\ 76.6\%), yielding approximately $2.3\times$ sample efficiency. Qwen3-32B is trained on 8$\times$A100 80\,GB (IOP-GSPO approximately 48 vs.\ GSPO approximately 46 GPU hours); Qwen3-Next is trained on 16$\times$A100 80\,GB (approximately 72 vs.\ 68 GPU hours). At both scales, the additional GPU hour overhead is $<$6\%, while performance improvement reaches 3.7--7.8\%, representing a favorable cost-performance trade-off.

\subsection{Inference Settings}
\label{app:inference-settings}

Both Qwen3-32B and Qwen3-Next use Temperature\,=\,0.6, TopP\,=\,0.95, TopK\,=\,20, MinP\,=\,0. All methods are compared under identical decoding settings to ensure that differences arise solely from the training method. All evaluations use 32 independent samples to compute avg@32.


\begin{thebibliography}{31}
\providecommand{\natexlab}[1]{#1}
\providecommand{\url}[1]{\texttt{#1}}
\expandafter\ifx\csname urlstyle\endcsname\relax
  \providecommand{\doi}[1]{doi: #1}\else
  \providecommand{\doi}{doi: \begingroup \urlstyle{rm}\Url}\fi

\bibitem[Ahmad et~al.(2025)Ahmad, Narenthiran, Majumdar, Ficek, Jain, Huang,
  Noroozi, and Ginsburg]{ahmad2025opencodereasoning}
Wasi~Uddin Ahmad, Sean Narenthiran, Somshubra Majumdar, Aleksander Ficek,
  Siddhartha Jain, Jocelyn Huang, Vahid Noroozi, and Boris Ginsburg.
\newblock Opencodereasoning: Advancing data distillation for competitive
  coding.
\newblock In \emph{Second Conference on Language Modeling}, 2025.
\newblock URL \url{https://openreview.net/forum?id=aykM7KUVJZ}.

\bibitem[Balunovic et~al.(2026)Balunovic, Dekoninck, Petrov, Jovanovi{\'c}, and
  Vechev]{balunovic2026matharena}
Mislav Balunovic, Jasper Dekoninck, Ivo Petrov, Nikola Jovanovi{\'c}, and
  Martin Vechev.
\newblock Matharena: Evaluating {LLM}s on uncontaminated math competitions.
\newblock In \emph{The Thirty-ninth Annual Conference on Neural Information
  Processing Systems Datasets and Benchmarks Track}, 2026.
\newblock URL \url{https://openreview.net/forum?id=y0zL9IZxZ7}.

\bibitem[Cui et~al.(2025)Cui, Yuan, Wang, Wang, Zhang, Chen, Li, He, Fan, Yu,
  Xu, Chen, Yuan, Chen, Zhang, Lv, Wang, Yao, Han, Peng, Cheng, Liu, Sun, Zhou,
  and Ding]{cui2025processreinforcementimplicitrewards}
Ganqu Cui, Lifan Yuan, Zefan Wang, Hanbin Wang, Yuchen Zhang, Jiacheng Chen,
  Wendi Li, Bingxiang He, Yuchen Fan, Tianyu Yu, Qixin Xu, Weize Chen, Jiarui
  Yuan, Huayu Chen, Kaiyan Zhang, Xingtai Lv, Shuo Wang, Yuan Yao, Xu~Han, Hao
  Peng, Yu~Cheng, Zhiyuan Liu, Maosong Sun, Bowen Zhou, and Ning Ding.
\newblock Process reinforcement through implicit rewards, 2025.
\newblock URL \url{https://arxiv.org/abs/2502.01456}.

\bibitem[Guo et~al.(2025)Guo, Yang, Zhang, Song, Wang, Zhu, Xu, Zhang, Ma, Bi,
  Zhang, Yu, Wu, Wu, Gou, Shao, Li, Gao, Liu, Xue, Wang, Wu, Feng, Lu, Zhao,
  Deng, Ruan, Dai, Chen, Ji, Li, Lin, Dai, Luo, Hao, Chen, Li, Zhang, Xu, Ding,
  Gao, Qu, Li, Guo, Li, Chen, Yuan, Tu, Qiu, Li, Cai, Ni, Liang, Chen, Dong,
  Hu, You, Gao, Guan, Huang, Yu, Wang, Zhang, Zhao, Wang, Zhang, Xu, Xia,
  Zhang, Zhang, Tang, Zhou, Li, Wang, Li, Tian, Huang, Zhang, Wang, Chen, Du,
  Ge, Zhang, Pan, Wang, Chen, Jin, Chen, Lu, Zhou, Chen, Ye, Wang, Yu, Zhou,
  Pan, Li, Zhou, Wu, Yun, Pei, Sun, Wang, Zeng, Liu, Liang, Gao, Yu, Zhang,
  Xiao, An, Liu, Wang, Chen, Nie, Cheng, Liu, Xie, Liu, Yang, Li, Su, Lin, Li,
  Jin, Shen, Chen, Sun, Wang, Song, Zhou, Wang, Shan, Li, Wang, Wei, Zhang, Xu,
  Li, Zhao, Sun, Wang, Yu, Zhang, Shi, Xiong, He, Piao, Wang, Tan, Ma, Liu,
  Guo, Ou, Wang, Gong, Zou, He, Xiong, Luo, You, Liu, Zhou, Zhu, Huang, Li,
  Zheng, Zhu, Ma, Tang, Zha, Yan, Ren, Ren, Sha, Fu, Xu, Xie, Zhang, Hao, Ma,
  Yan, Wu, Gu, Zhu, Liu, Li, Xie, Song, Pan, Huang, Xu, Zhang, and
  Zhang]{Guo_2025}
Daya Guo, Dejian Yang, Haowei Zhang, Junxiao Song, Peiyi Wang, Qihao Zhu,
  Runxin Xu, Ruoyu Zhang, Shirong Ma, Xiao Bi, Xiaokang Zhang, Xingkai Yu,
  Yu~Wu, Z.~F. Wu, Zhibin Gou, Zhihong Shao, Zhuoshu Li, Ziyi Gao, Aixin Liu,
  Bing Xue, Bingxuan Wang, Bochao Wu, Bei Feng, Chengda Lu, Chenggang Zhao,
  Chengqi Deng, Chong Ruan, Damai Dai, Deli Chen, Dongjie Ji, Erhang Li,
  Fangyun Lin, Fucong Dai, Fuli Luo, Guangbo Hao, Guanting Chen, Guowei Li,
  H.~Zhang, Hanwei Xu, Honghui Ding, Huazuo Gao, Hui Qu, Hui Li, Jianzhong Guo,
  Jiashi Li, Jingchang Chen, Jingyang Yuan, Jinhao Tu, Junjie Qiu, Junlong Li,
  J.~L. Cai, Jiaqi Ni, Jian Liang, Jin Chen, Kai Dong, Kai Hu, Kaichao You,
  Kaige Gao, Kang Guan, Kexin Huang, Kuai Yu, Lean Wang, Lecong Zhang, Liang
  Zhao, Litong Wang, Liyue Zhang, Lei Xu, Leyi Xia, Mingchuan Zhang, Minghua
  Zhang, Minghui Tang, Mingxu Zhou, Meng Li, Miaojun Wang, Mingming Li, Ning
  Tian, Panpan Huang, Peng Zhang, Qiancheng Wang, Qinyu Chen, Qiushi Du, Ruiqi
  Ge, Ruisong Zhang, Ruizhe Pan, Runji Wang, R.~J. Chen, R.~L. Jin, Ruyi Chen,
  Shanghao Lu, Shangyan Zhou, Shanhuang Chen, Shengfeng Ye, Shiyu Wang,
  Shuiping Yu, Shunfeng Zhou, Shuting Pan, S.~S. Li, Shuang Zhou, Shaoqing Wu,
  Tao Yun, Tian Pei, Tianyu Sun, T.~Wang, Wangding Zeng, Wen Liu, Wenfeng
  Liang, Wenjun Gao, Wenqin Yu, Wentao Zhang, W.~L. Xiao, Wei An, Xiaodong Liu,
  Xiaohan Wang, Xiaokang Chen, Xiaotao Nie, Xin Cheng, Xin Liu, Xin Xie,
  Xingchao Liu, Xinyu Yang, Xinyuan Li, Xuecheng Su, Xuheng Lin, X.~Q. Li,
  Xiangyue Jin, Xiaojin Shen, Xiaosha Chen, Xiaowen Sun, Xiaoxiang Wang, Xinnan
  Song, Xinyi Zhou, Xianzu Wang, Xinxia Shan, Y.~K. Li, Y.~Q. Wang, Y.~X. Wei,
  Yang Zhang, Yanhong Xu, Yao Li, Yao Zhao, Yaofeng Sun, Yaohui Wang, Yi~Yu,
  Yichao Zhang, Yifan Shi, Yiliang Xiong, Ying He, Yishi Piao, Yisong Wang,
  Yixuan Tan, Yiyang Ma, Yiyuan Liu, Yongqiang Guo, Yuan Ou, Yuduan Wang, Yue
  Gong, Yuheng Zou, Yujia He, Yunfan Xiong, Yuxiang Luo, Yuxiang You, Yuxuan
  Liu, Yuyang Zhou, Y.~X. Zhu, Yanping Huang, Yaohui Li, Yi~Zheng, Yuchen Zhu,
  Yunxian Ma, Ying Tang, Yukun Zha, Yuting Yan, Z.~Z. Ren, Zehui Ren, Zhangli
  Sha, Zhe Fu, Zhean Xu, Zhenda Xie, Zhengyan Zhang, Zhewen Hao, Zhicheng Ma,
  Zhigang Yan, Zhiyu Wu, Zihui Gu, Zijia Zhu, Zijun Liu, Zilin Li, Ziwei Xie,
  Ziyang Song, Zizheng Pan, Zhen Huang, Zhipeng Xu, Zhongyu Zhang, and Zhen
  Zhang.
\newblock Deepseek-r1 incentivizes reasoning in llms through reinforcement
  learning.
\newblock \emph{Nature}, 645\penalty0 (8081):\penalty0 633–638, September
  2025.
\newblock ISSN 1476-4687.
\newblock \doi{10.1038/s41586-025-09422-z}.
\newblock URL \url{http://dx.doi.org/10.1038/s41586-025-09422-z}.

\bibitem[Havrilla et~al.(2024)Havrilla, Du, Raparthy, Nalmpantis, Dwivedi-Yu,
  Zhuravinskyi, Hambro, Sukhbaatar, and
  Raileanu]{havrilla2024teachinglargelanguagemodels}
Alex Havrilla, Yuqing Du, Sharath~Chandra Raparthy, Christoforos Nalmpantis,
  Jane Dwivedi-Yu, Maksym Zhuravinskyi, Eric Hambro, Sainbayar Sukhbaatar, and
  Roberta Raileanu.
\newblock Teaching large language models to reason with reinforcement learning,
  2024.
\newblock URL \url{https://arxiv.org/abs/2403.04642}.

\bibitem[He et~al.(2026)He, Liang, Xu, Liu, Chen, Wang, Song, Yu, Liang, Wang,
  Zhang, Wang, Tu, Mi, and Yu]{he2026deepmathk}
Zhiwei He, Tian Liang, Jiahao Xu, Qiuzhi Liu, Xingyu Chen, Yue Wang, Linfeng
  Song, Dian Yu, Zhenwen Liang, Wenxuan Wang, Zhuosheng Zhang, Rui Wang,
  Zhaopeng Tu, Haitao Mi, and Dong Yu.
\newblock Deepmath-103k: A large-scale, challenging, decontaminated, and
  verifiable mathematical dataset for advancing reasoning.
\newblock In \emph{The Fourteenth International Conference on Learning
  Representations}, 2026.
\newblock URL \url{https://openreview.net/forum?id=kHB5Te5IWm}.

\bibitem[Jain et~al.(2025)Jain, Han, Gu, Li, Yan, Zhang, Wang, Solar-Lezama,
  Sen, and Stoica]{jain2025livecodebench}
Naman Jain, King Han, Alex Gu, Wen-Ding Li, Fanjia Yan, Tianjun Zhang, Sida
  Wang, Armando Solar-Lezama, Koushik Sen, and Ion Stoica.
\newblock Livecodebench: Holistic and contamination free evaluation of large
  language models for code.
\newblock In \emph{The Thirteenth International Conference on Learning
  Representations}, 2025.
\newblock URL \url{https://openreview.net/forum?id=chfJJYC3iL}.

\bibitem[Khalifa et~al.(2026)Khalifa, Agarwal, Logeswaran, Kim, Peng, Lee, Lee,
  and Wang]{khalifa2026process}
Muhammad Khalifa, Rishabh Agarwal, Lajanugen Logeswaran, Jaekyeom Kim, Hao
  Peng, Moontae Lee, Honglak Lee, and Lu~Wang.
\newblock Process reward models that think.
\newblock \emph{Transactions on Machine Learning Research}, 2026.
\newblock ISSN 2835-8856.
\newblock URL \url{https://openreview.net/forum?id=FPVCb0WMuN}.
\newblock J2C Certification.

\bibitem[Kumar et~al.(2025)Kumar, Zhuang, Agarwal, Su, Co-Reyes, Singh, Baumli,
  Iqbal, Bishop, Roelofs, Zhang, McKinney, Shrivastava, Paduraru, Tucker,
  Precup, Behbahani, and Faust]{kumar2025training}
Aviral Kumar, Vincent Zhuang, Rishabh Agarwal, Yi~Su, John~D Co-Reyes, Avi
  Singh, Kate Baumli, Shariq Iqbal, Colton Bishop, Rebecca Roelofs, Lei~M
  Zhang, Kay McKinney, Disha Shrivastava, Cosmin Paduraru, George Tucker, Doina
  Precup, Feryal Behbahani, and Aleksandra Faust.
\newblock Training language models to self-correct via reinforcement learning.
\newblock In \emph{The Thirteenth International Conference on Learning
  Representations}, 2025.
\newblock URL \url{https://openreview.net/forum?id=CjwERcAU7w}.

\bibitem[Lee et~al.(2025)Lee, Oh, Tack, Kim, and Shin]{lee2025revise}
Hyunseok Lee, Seunghyuk Oh, Jihoon Tack, Jaehyung Kim, and Jinwoo Shin.
\newblock Re{VISE}: Learning to refine at test-time via intrinsic
  self-verification.
\newblock In \emph{Workshop on Reasoning and Planning for Large Language
  Models}, 2025.
\newblock URL \url{https://openreview.net/forum?id=0lGvQDPKwh}.

\bibitem[Liang et~al.(2026)Liang, Zhu, Ge, Yang, Shen, Zheng, and
  Guo]{liang2026learningirrecoverableerrorlocalizedpolicy}
Qiao Liang, Yuke Zhu, Chao Ge, Lei Yang, Ying Shen, Bo~Zheng, and Sheng Guo.
\newblock Learning from the irrecoverable: Error-localized policy optimization
  for tool-integrated llm reasoning, 2026.
\newblock URL \url{https://arxiv.org/abs/2602.09598}.

\bibitem[Lightman et~al.(2024)Lightman, Kosaraju, Burda, Edwards, Baker, Lee,
  Leike, Schulman, Sutskever, and Cobbe]{lightman2024lets}
Hunter Lightman, Vineet Kosaraju, Yuri Burda, Harrison Edwards, Bowen Baker,
  Teddy Lee, Jan Leike, John Schulman, Ilya Sutskever, and Karl Cobbe.
\newblock Let's verify step by step.
\newblock In \emph{The Twelfth International Conference on Learning
  Representations}, 2024.
\newblock URL \url{https://openreview.net/forum?id=v8L0pN6EOi}.

\bibitem[Liu et~al.(2026)Liu, Yu, Lu, Zhou, Liu, Liang, Mi, Wei, and
  Yu]{liu2026savegoodprefixprecise}
Haolin Liu, Dian Yu, Sidi Lu, Yujun Zhou, Rui Liu, Zhenwen Liang, Haitao Mi,
  Chen-Yu Wei, and Dong Yu.
\newblock Save the good prefix: Precise error penalization via
  process-supervised rl to enhance llm reasoning, 2026.
\newblock URL \url{https://arxiv.org/abs/2601.18984}.

\bibitem[Luo et~al.(2024)Luo, Liu, Liu, Phatale, Guo, Lara, Li, Shu, Zhu, Meng,
  Sun, and Rastogi]{luo2024improvemathematicalreasoninglanguage}
Liangchen Luo, Yinxiao Liu, Rosanne Liu, Samrat Phatale, Meiqi Guo, Harsh Lara,
  Yunxuan Li, Lei Shu, Yun Zhu, Lei Meng, Jiao Sun, and Abhinav Rastogi.
\newblock Improve mathematical reasoning in language models by automated
  process supervision, 2024.
\newblock URL \url{https://arxiv.org/abs/2406.06592}.

\bibitem[Ma et~al.(2025)Ma, Wang, Liu, Liu, Chen, Zhang, Zhou, Du, and
  Li]{ma-etal-2025-s2r}
Ruotian Ma, Peisong Wang, Cheng Liu, Xingyan Liu, Jiaqi Chen, Bang Zhang, Xin
  Zhou, Nan Du, and Jia Li.
\newblock {S}$^2${R}: Teaching {LLM}s to self-verify and self-correct via
  reinforcement learning.
\newblock In Wanxiang Che, Joyce Nabende, Ekaterina Shutova, and Mohammad~Taher
  Pilehvar (eds.), \emph{Proceedings of the 63rd Annual Meeting of the
  Association for Computational Linguistics (Volume 1: Long Papers)}, pp.\
  22632--22654, Vienna, Austria, July 2025. Association for Computational
  Linguistics.
\newblock ISBN 979-8-89176-251-0.
\newblock \doi{10.18653/v1/2025.acl-long.1104}.
\newblock URL \url{https://aclanthology.org/2025.acl-long.1104/}.

\bibitem[Madaan et~al.(2023)Madaan, Tandon, Gupta, Hallinan, Gao, Wiegreffe,
  Alon, Dziri, Prabhumoye, Yang, Gupta, Majumder, Hermann, Welleck,
  Yazdanbakhsh, and Clark]{NEURIPS2023_91edff07}
Aman Madaan, Niket Tandon, Prakhar Gupta, Skyler Hallinan, Luyu Gao, Sarah
  Wiegreffe, Uri Alon, Nouha Dziri, Shrimai Prabhumoye, Yiming Yang, Shashank
  Gupta, Bodhisattwa~Prasad Majumder, Katherine Hermann, Sean Welleck, Amir
  Yazdanbakhsh, and Peter Clark.
\newblock Self-refine: Iterative refinement with self-feedback.
\newblock In A.~Oh, T.~Naumann, A.~Globerson, K.~Saenko, M.~Hardt, and
  S.~Levine (eds.), \emph{Advances in Neural Information Processing Systems},
  volume~36, pp.\  46534--46594. Curran Associates, Inc., 2023.
\newblock URL
  \url{https://proceedings.neurips.cc/paper_files/paper/2023/file/91edff07232fb1b55a505a9e9f6c0ff3-Paper-Conference.pdf}.

\bibitem[{Mathematical Association of America}(2024)]{maa_aime2025}
{Mathematical Association of America}.
\newblock 2024-25 {AIME} thresholds are available.
\newblock \url{https://maa.org/aime-thresholds-are-available/}, 2024.
\newblock Updated January 6, 2025. Accessed March 21, 2026.

\bibitem[Nie et~al.(2026)Nie, Ding, Zhang, Yu, Yang, Chen, Yin, Sun, Wu, and
  Liu]{nie2026attnpoattentionguidedprocesssupervision}
Shuaiyi Nie, Siyu Ding, Wenyuan Zhang, Linhao Yu, Tianmeng Yang, Yao Chen,
  Weichong Yin, Yu~Sun, Hua Wu, and Tingwen Liu.
\newblock Attnpo: Attention-guided process supervision for efficient reasoning,
  2026.
\newblock URL \url{https://arxiv.org/abs/2602.09953}.

\bibitem[{OpenAI}(2024)]{openai2024o1}
{OpenAI}.
\newblock Learning to reason with {LLMs}.
\newblock \url{https://openai.com/index/learning-to-reason-with-llms/}, 2024.
\newblock OpenAI release, September 12, 2024.

\bibitem[OpenAI(2025)]{openai2025gptoss120bgptoss20bmodel}
OpenAI.
\newblock gpt-oss-120b \& gpt-oss-20b model card, 2025.
\newblock URL \url{https://arxiv.org/abs/2508.10925}.

\bibitem[Pronesti et~al.(2026)Pronesti, Belz, and
  Hou]{pronesti2026outcomeverificationverifiableprocess}
Massimiliano Pronesti, Anya Belz, and Yufang Hou.
\newblock Beyond outcome verification: Verifiable process reward models for
  structured reasoning, 2026.
\newblock URL \url{https://arxiv.org/abs/2601.17223}.

\bibitem[Rafailov et~al.(2023)Rafailov, Sharma, Mitchell, Manning, Ermon, and
  Finn]{NEURIPS2023_a85b405e}
Rafael Rafailov, Archit Sharma, Eric Mitchell, Christopher~D Manning, Stefano
  Ermon, and Chelsea Finn.
\newblock Direct preference optimization: Your language model is secretly a
  reward model.
\newblock In A.~Oh, T.~Naumann, A.~Globerson, K.~Saenko, M.~Hardt, and
  S.~Levine (eds.), \emph{Advances in Neural Information Processing Systems},
  volume~36, pp.\  53728--53741. Curran Associates, Inc., 2023.
\newblock URL
  \url{https://proceedings.neurips.cc/paper_files/paper/2023/file/a85b405ed65c6477a4fe8302b5e06ce7-Paper-Conference.pdf}.

\bibitem[Shao et~al.(2024)Shao, Wang, Zhu, Xu, Song, Bi, Zhang, Zhang, Li, Wu,
  and Guo]{shao2024deepseekmathpushinglimitsmathematical}
Zhihong Shao, Peiyi Wang, Qihao Zhu, Runxin Xu, Junxiao Song, Xiao Bi, Haowei
  Zhang, Mingchuan Zhang, Y.~K. Li, Y.~Wu, and Daya Guo.
\newblock Deepseekmath: Pushing the limits of mathematical reasoning in open
  language models, 2024.
\newblock URL \url{https://arxiv.org/abs/2402.03300}.

\bibitem[Shinn et~al.(2023)Shinn, Cassano, Gopinath, Narasimhan, and
  Yao]{NEURIPS2023_1b44b878}
Noah Shinn, Federico Cassano, Ashwin Gopinath, Karthik Narasimhan, and Shunyu
  Yao.
\newblock Reflexion: language agents with verbal reinforcement learning.
\newblock In A.~Oh, T.~Naumann, A.~Globerson, K.~Saenko, M.~Hardt, and
  S.~Levine (eds.), \emph{Advances in Neural Information Processing Systems},
  volume~36, pp.\  8634--8652. Curran Associates, Inc., 2023.
\newblock URL
  \url{https://proceedings.neurips.cc/paper_files/paper/2023/file/1b44b878bb782e6954cd888628510e90-Paper-Conference.pdf}.

\bibitem[Wang et~al.(2024)Wang, Li, Shao, Xu, Dai, Li, Chen, Wu, and
  Sui]{wang-etal-2024-math}
Peiyi Wang, Lei Li, Zhihong Shao, Runxin Xu, Damai Dai, Yifei Li, Deli Chen,
  Yu~Wu, and Zhifang Sui.
\newblock Math-shepherd: Verify and reinforce {LLM}s step-by-step without human
  annotations.
\newblock In Lun-Wei Ku, Andre Martins, and Vivek Srikumar (eds.),
  \emph{Proceedings of the 62nd Annual Meeting of the Association for
  Computational Linguistics (Volume 1: Long Papers)}, pp.\  9426--9439,
  Bangkok, Thailand, August 2024. Association for Computational Linguistics.
\newblock \doi{10.18653/v1/2024.acl-long.510}.
\newblock URL \url{https://aclanthology.org/2024.acl-long.510/}.

\bibitem[Wen et~al.(2026)Wen, Liu, Zheng, Ye, Wu, Wang, Xu, Liang, Li, Miao,
  Bian, and Yang]{wen2026reinforcement}
Xumeng Wen, Zihan Liu, Shun Zheng, Shengyu Ye, Zhirong Wu, Yang Wang, Zhijian
  Xu, Xiao Liang, Junjie Li, Ziming Miao, Jiang Bian, and Mao Yang.
\newblock Reinforcement learning with verifiable rewards implicitly
  incentivizes correct reasoning in base {LLM}s.
\newblock In \emph{The Fourteenth International Conference on Learning
  Representations}, 2026.
\newblock URL \url{https://openreview.net/forum?id=jGbRWwIidy}.

\bibitem[Xiong et~al.(2025)Xiong, Zhang, Ye, Chen, Jiang, and
  Zhang]{xiong2025selfrewardingcorrectionmathematicalreasoning}
Wei Xiong, Hanning Zhang, Chenlu Ye, Lichang Chen, Nan Jiang, and Tong Zhang.
\newblock Self-rewarding correction for mathematical reasoning, 2025.
\newblock URL \url{https://arxiv.org/abs/2502.19613}.

\bibitem[Yang et~al.(2025)Yang, He, Shi, Deng, Li, Yin, and
  Jiang]{yang-etal-2025-beyond-first}
Zhaohui Yang, Chenghua He, Xiaowen Shi, Shihong Deng, Linjing Li, Qiyue Yin,
  and Daxin Jiang.
\newblock Beyond the first error: Process reward models for reflective
  mathematical reasoning.
\newblock In Christos Christodoulopoulos, Tanmoy Chakraborty, Carolyn Rose, and
  Violet Peng (eds.), \emph{Findings of the Association for Computational
  Linguistics: EMNLP 2025}, pp.\  4711--4728, Suzhou, China, November 2025.
  Association for Computational Linguistics.
\newblock ISBN 979-8-89176-335-7.
\newblock \doi{10.18653/v1/2025.findings-emnlp.253}.
\newblock URL \url{https://aclanthology.org/2025.findings-emnlp.253/}.

\bibitem[Yao et~al.(2026)Yao, Wang, and Zhang]{yao2026prlprocessrewardlearning}
Jiarui Yao, Ruida Wang, and Tong Zhang.
\newblock Prl: Process reward learning improves llms' reasoning ability and
  broadens the reasoning boundary, 2026.
\newblock URL \url{https://arxiv.org/abs/2601.10201}.

\bibitem[Zhang et~al.(2025)Zhang, Zheng, Wu, Zhang, Lin, Yu, Liu, Zhou, and
  Lin]{zhang-etal-2025-lessons}
Zhenru Zhang, Chujie Zheng, Yangzhen Wu, Beichen Zhang, Runji Lin, Bowen Yu,
  Dayiheng Liu, Jingren Zhou, and Junyang Lin.
\newblock The lessons of developing process reward models in mathematical
  reasoning.
\newblock In Wanxiang Che, Joyce Nabende, Ekaterina Shutova, and Mohammad~Taher
  Pilehvar (eds.), \emph{Findings of the Association for Computational
  Linguistics: ACL 2025}, pp.\  10495--10516, Vienna, Austria, July 2025.
  Association for Computational Linguistics.
\newblock ISBN 979-8-89176-256-5.
\newblock \doi{10.18653/v1/2025.findings-acl.547}.
\newblock URL \url{https://aclanthology.org/2025.findings-acl.547/}.

\bibitem[Zheng et~al.(2025)Zheng, Liu, Li, Chen, Yu, Gao, Dang, Liu, Men, Yang,
  Zhou, and Lin]{zheng2025groupsequencepolicyoptimization}
Chujie Zheng, Shixuan Liu, Mingze Li, Xiong-Hui Chen, Bowen Yu, Chang Gao, Kai
  Dang, Yuqiong Liu, Rui Men, An~Yang, Jingren Zhou, and Junyang Lin.
\newblock Group sequence policy optimization, 2025.
\newblock URL \url{https://arxiv.org/abs/2507.18071}.

\end{thebibliography}
\end{document}